# Boolean Equi-propagation
# for Optimized SAT Encoding

Amit Metodi[1], Michael Codish[1], Vitaly Lagoon[2], and Peter J. Stuckey[3]

[1] Department of Computer Science, Ben Gurion University of the Negev, Israel
[2] Cadence Design Systems, USA
[3] Department of Computer Science and Software Engineering, and
NICTA Victoria Laboratory, The University of Melbourne, Australia

**Abstract.** We present an approach to propagation based solving, Boolean equi-propagation, where constraints are modelled as propagators of information about equalities between Boolean literals. Propagation based solving applies this information as a form of partial evaluation resulting in optimized SAT encodings. We demonstrate for a variety of benchmarks that our approach results in smaller CNF encodings and leads to speed-ups in solving times.

## 1 Introduction

In recent years, Boolean SAT solving techniques have improved dramatically. Today's SAT solvers are considerably faster and able to manage far larger instances than yesterday's. Moreover, encoding and modeling techniques are better understood and increasingly innovative. SAT is currently applied to solve a wide variety of hard and practical combinatorial problems, often outperforming dedicated algorithms. The general idea is to encode a (typically, NP) hard problem instance, $P$, to a Boolean formula, $\varphi_P$, such that the solutions of $P$ correspond to the satisfying assignments of $\varphi_P$. Given an encoding from problem instances to Boolean formula, a SAT solver is then applied to solve the problem instances.

Tailgating the success of SAT technology are a variety of tools which can be applied to specify and then compile problem instances to corresponding SAT instances. Typically, a constraint based modelling language is introduced and used to model instances. Then encoding techniques are applied to compile constraints to the language of an underlying solver such as SAT, SMT, or others. Some examples follow: In [5], Cadoli and Schaerf introduce a logic-based specification language, called NP-SPEC, which allows one to specify combinatorial problems in a declarative way. At the core of their system is a compiler which translates specifications to CNF formula. Sugar [19], is a SAT-based constraint solver. To solve a finite domain linear constraint satisfaction problem it is first encoded to a CNF formula by Sugar, and then solved using the MiniSat solver [8]. MiniZinc [14], is a constraint modeling language which is compiled by a variety of solvers to the low-level target language FlatZinc. For example, fzntini [13] solves FlatZinc instances by encoding them as CNF. In [3], the authors present fzn2smt, a tool for solving FlatZinc instances by encoding them to SMT-LIB [2].

Taking the analogy with programming languages, we want to facilitate the process of providing a high-level description of how the (constraint) problem at hand is to be solved. Given such a description, a compiler can then provide a, low-level executable for the underlying machine. In our context, the low-level executable is a CNF formula, and the underlying machine, a SAT solver. One advantage in the use of such tools is that the user can easily experiment with a variety of modeling options without the need to tediously encode each as a CNF formula. Another advantage is that optimizations can be handled in the compiler, and not each time from scratch.

A major obstacle when solving combinatorial problems using SAT technology is that CNF encodings do not maintain the context of the bits specified in the constraint model. For example, the information that certain bits in the CNF encoding originate from a vector representing an integer value in the constraint model, is lost. Consequently, our ability to apply optimizations related to the original vector diminish.

This paper takes a new approach, defining the notion of an equi-propagation solver. Prior to SAT encoding, constraints are viewed as propagators of information about equalities between Boolean literals and constants. We repeatedly extract such equalities and then factor them out of the original constraint problem. We demonstrate that this significantly reduces the size of the constraint problem and the subsequent SAT solving time. A novel and efficient implementation of equi-propagation using binary decision diagrams (BDD's) [4] is described. Drawing on the programming languages analogy, we contribute an optimizing compiler for SAT encoding. Equi-propagation and partial evaluation facilitate optimization of the constraint model. This, fast (polynomial-time) optimization phase is followed by the more costly (exponential-time) SAT solving phase.

## 2 Overview

Constraints in our modelling language are viewed as a Boolean functions about the underlying bit representation for finite domain integer variables and other Boolean variables. We focus on a unary representation, the *order encoding*, for integer variables. Consider three (standard) constructs in the modelling language.

$$\boxed{1}\ \texttt{unary}_n(X, [a, b]) \qquad \boxed{2}\ \texttt{diff}(X_1, X_2) \qquad \boxed{3}\ \texttt{allDiff}([X_1, \ldots, X_n])$$

A constraint $\texttt{unary}_n(X, [a, b])$ where $0 \leq a \leq b \leq n$ specifies a finite domain integer variable $X = \langle x_1, \ldots, x_n \rangle$, represented in $n$ bits, which takes values in the interval $[a, b]$. We denote by $dom(X)$ the finite set of values that variable $X$ can take. Initially, $dom(X) = \{a, \ldots, b\}$. When clear from the context, we drop $n$ from the notation. A constraint, $\texttt{diff}(X_1, X_2)$, specifies that integer variables (bit vectors) $X_1$ and $X_2$ take different values from their respective domains. The third construct, $\texttt{allDiff}([X_1, \ldots, X_m])$, specifies that integer variables $[X_1, \ldots, X_m]$ all take different values from their respective domains. The argument of this constraint is a list of bit vectors. We denote $dom([X_1, \ldots, X_m]) = \cup \{\, dom(X_i) \,|\, 1 \leq i \leq m \,\}$.

In the *order encoding* (see e.g. [6, 1]), the bit vector representation of integer variable $X = \langle x_1, \ldots, x_n \rangle$ constitutes a monotonic decreasing sequence. For example, the value 3 in 5 bits is represented as $\langle 1, 1, 1, 0, 0 \rangle$. The bit $x_i$ (for $1 \leq i \leq n$) is interpreted as the statement $X \geq i$. Throughout the paper, for a bit vector $X = \langle x_1, \ldots, x_n \rangle$ representing an integer in the order-encoding, we assume implicit bits $x_0 = 1$ and $x_{n+1} = 0$, and denote $X(i) = x_i$ for $0 \leq i \leq n+1$. The order encoding is also used in Sugar [19].

The Boolean functions corresponding to constraints $\boxed{1}$ — $\boxed{3}$ are as follows (where $1 \leq a \leq b \leq n$):

$$\begin{aligned}
\texttt{unary}(\langle x_1, \ldots, x_n \rangle, [a,b]) &= \bigwedge_{i=1}^{n} (x_{i-1} \leftarrow x_i) \wedge x_a \wedge \neg x_{b+1} \\
\texttt{diff}(\langle x_1, \ldots, x_n \rangle, \langle y_1, \ldots, y_n \rangle) &= \bigvee_{i=1}^{n} (x_i \texttt{ xor } y_i) \\
\texttt{allDiff}([X_1, \ldots, X_m]) &= \bigwedge_{1 \leq i < j \leq m} \texttt{diff}(X_i, X_j)
\end{aligned} \quad (1)$$

For constraint $c$ with integer variable arguments, we denote by $c_u$ the conjunction of $c$ with the statement that its arguments are represented in the order-encoding. For example, $\texttt{diff}_u(X, Y) = \texttt{diff}(X, Y) \wedge \texttt{unary}_n(X, [0, n]) \wedge \texttt{unary}_n(Y, [0, n])$.

An important property of a Boolean representation for finite domain integers is the ability to represent changes in the set of values a variable can take. It is well-known that the order-encoding facilitates the propagation of bounds. Consider an integer variable $X = \langle x_1, \ldots, x_n \rangle$ with values in the interval $[0, n]$. To restrict $X$ to take values in the range $[a, b]$ (for $1 \leq a \leq b \leq n$), it is sufficient to assign $x_a = 1$ and $x_{b+1} = 0$ (if $b < n$). The variables $x_{a'}$ for $0 \geq a' > a$ and $b < b' \leq n$ are then determined true and false, respectively, by *unit propagation*. For example, given $X = \langle x_1, \ldots, x_9 \rangle$, assigning $x_3 = 1$ and $x_6 = 0$ propagates to give $X = \langle 1, 1, 1, x_4, x_5, 0, 0, 0, 0 \rangle$, signifying that $dom(X) \subseteq \{3, \ldots, 5\}$.

A lesser known property of the order-encoding is its ability to specify that a variable cannot take a specific value $0 \leq v \leq n$ in its domain by equating two variables: $x_v = x_{v+1}$. This indicates that the order-encoding is well-suited not only to propagate lower and upper bounds, but also to represent integer variables with an arbitrary, finite set, domain. For example, for $X = \langle x_1, \ldots, x_9 \rangle$, equating $x_2 = x_3$ imposes that $X \neq 2$. Likewise $x_5 = x_6$ and $x_7 = x_8$ impose that $X \neq 5$ and $X \neq 7$. Applying these equalities to $X$ gives, $X = \langle x_1, x_2, x_2, x_4, x_5, x_5, x_7, x_7, x_9 \rangle$, signifying that $dom(X) = \{0, 1, 3, 4, 6, 8, 9\}$.

The idea in this paper is to simplify constraints, prior to their encoding to CNF, using a technique we call equi-propagation. We distinguish between *low-level constraints*, such as $\texttt{unary}(X, [a, b])$ and $\texttt{diff}(X_1, X_2)$, which are about a fixed number (one and two) of integer variables, and *high-level constraints*, such as $\texttt{allDiff}([X_1, \ldots, X_m])$. Low-level constraints are simplified and then encoded directly to CNF, while high-level constraints are simplified and then decomposed to low-level constraints. We consider three types of simplification rules. To illustrate these, consider the constraint $\texttt{diff}(X, Y)$ where $X = \langle x_1, x_2, x_3, x_4 \rangle$ and $Y = \langle y_1, y_2, y_3, y_4 \rangle$ are unary variables in the order-encoding.

**(1)** *equi-propagation*, where we propagate information about equalities between Boolean literals and constants. For example, given equalities s.t. $Y = \langle 1, 1, 0, 0 \rangle$ we propagate that $(x_2 = x_3)$ because $\mathtt{diff}_u(\langle x_1, x_2, x_3, x_4\rangle, \langle 1,1,0,0\rangle) \models (x_2 = x_3)$. ($X$ is in the order-encoding, so $x_2 \geq x_3$, and $x_2 \leq x_3$ as otherwise $x_2 = 1$ and $x_3 = 0$ which implies that $X = \langle 1,1,0,0\rangle$, contradicting $\mathtt{diff}(X, Y)$). When we detect such equalities, we apply them to simplify the constraints in a model.

**(2)** *redundant constraint elimination*, where we discover that, due to equalities, a constraint is redundant. For example, when $Y = \langle 1,1,0,0\rangle$ and $x_2 = x_3$, the constraint $\mathtt{diff}(X,Y)$ is redundant because $\mathtt{unary}(\langle x_1,x_2,x_3,x_4\rangle, [0,4]) \models \mathtt{diff}_u(\langle x_1,x_2,x_3,x_4\rangle, \langle 1,1,0,0\rangle)$.

**(3)** *constraint restriction*, where we discover that some bits in a constraint $c$ are "dont-cares" and project $c$ to the remaining variables. For example, when $x_1 = 1$ and $x_2 = 1$ then $y_1$ is a don't care and $\mathtt{diff}(X,Y)$ is equivalent to $\mathtt{diff}(X', Y')$ where $X' = \langle x_2, x_3, x_4\rangle$ and $Y' = \langle y_2, y_3, y_4\rangle$. To see why $\mathtt{unary}(X, [0,4]) \wedge \mathtt{unary}(Y, [0,4]) \wedge x_1 = 1 \wedge x_2 = 1 \models \mathtt{diff}(X,Y) \leftrightarrow \mathtt{diff}(X', Y')$, consider that if $y_1 = 0$ then also $y_2 = 0$ and both constraints are true, and if $y_1 = 1$ then $x_1 \mathtt{\ xor\ } y_1 = \mathit{false}$ and $\mathtt{diff}(X,Y) \leftrightarrow \mathtt{diff}(X', Y')$ follows.

In addition to simplification rules, we apply *decomposition rules* to high-level constraints. For example, an $\mathtt{allDiff}$ constraint decomposes naturally to a set of constituent $\mathtt{diff}$ constraints. The rule we apply to decompose $\mathtt{allDiff}$ constraints is as follows:

$$\mathtt{allDiff}([U_1, \ldots, U_m]) \mapsto \{ \mathtt{diff}(U_i, U_j) \mid 1 \leq i < j \leq m \}, \qquad (2)$$
$$\mathtt{permutation}_\#([U_1, \ldots, U_m])$$

where $\mathtt{permutation}_\#$ is a redundant constraint.[4] Its role is to introduce redundant clauses to accelerate SAT solving for the special case when the $\mathtt{allDiff}$ constraint specifies a permutation ($m$ variables taking $m$ different values). By delaying the special treatment of $\mathtt{allDiff}$ constraints which specify permutations we can often detect more permutations than prior to constraint simplification. The precise specification of the $\mathtt{permutation}_\#$ constraint is given in Section 4.

## 3 Boolean Equi-Propagation

Let $\mathcal{B}$ be a set of Boolean variables. A *literal* is a Boolean variable $b \in \mathcal{B}$ or its negation $\neg b$. The negation of a literal $\ell$, denoted $\neg\ell$, is defined as $\neg b$ if $\ell = b$ and as $b$ if $\ell = \neg b$. The Boolean constants 1 and 0 represent *true* and *false*, respectively. The set of literals is denoted $\mathcal{L}$ and $\mathcal{L}_{0,1} = \mathcal{L} \cup \{0, 1\}$.

An *assignment*, $A$, is a partial mapping from Boolean variables to constants, often viewed as the set of literals: $\{ b \mid A(b) = 1 \} \cup \{ \neg b \mid A(b) = 0 \}$. For a formula $\varphi$ and $b \in \mathcal{B}$, we denote by $\varphi[b]$ (likewise $\varphi[\neg b]$) the formula obtained by substituting all occurrences of $b \in \mathcal{B}$ in $\varphi$ by *true* (*false*). This notation extends in the natural way for sets of literals. We say that $A$ satisfies $\varphi$ if $\varphi[A]$ evaluates to *true*. A *Boolean Satisfiability (SAT) problem* consists of a Boolean formula $\varphi$ and determines if there exists an assignment which satisfies $\varphi$. The set of (free) Boolean variables that appear in a Boolean formula $\varphi$ is denoted $vars(\varphi)$.

---

[4] the symbol $_\#$ in the name of a constraint indicates that it is redundant.

A *Boolean equality* is a constraint $\ell = \ell'$ where $\ell, \ell' \in \mathcal{L}_{0,1}$. A *(Boolean) equi-formula* $E$ is a set of Boolean equalities understood as a conjunction. The set of equi-formulae is denoted $\mathcal{E}$.

**Equi-propagation** is the process of inferring new equational consequences from the constraints of a model and existing equational information. An *equi-propagator* for Boolean formula $\varphi$ is an extensive function $\mu_\varphi : \mathcal{E} \to \mathcal{E}$ (namely, s.t. $\mu_\varphi(E) \to E$) defined s.t. $\wedge \{ e \in \mathcal{E} \,|\, \varphi \wedge E \models e \} \to \mu_\varphi(E)$. That is, a conjunction of Boolean equalities, at least as strong as $E$, made true by $\varphi \wedge E$. We say that equi-propagator $\mu_\varphi$ is complete if for all equi-formula $E$, $\mu_\varphi(E) \leftrightarrow \{ e \in \mathcal{E} \,|\, \varphi \wedge E \models e \}$. We denote a complete equi-propagator for $\varphi$ as $\hat{\mu}_\varphi$.

*Example 1.* Let $X = \langle x_1, x_2, x_3, x_4 \rangle$ and $Y = \langle y_1, y_2, y_3, y_4 \rangle$ and consider $E_1 = \{ y_1 = 1,\ y_2 = 1,\ y_3 = 0,\ y_4 = 0 \}$ and $E_2 = \{ x_2 = \neg y_3,\ x_3 = \neg y_2 \}$. Then, $\hat{\mu}_{\texttt{diff}_u(X,Y)}(E_1) = E_1 \cup \{ x_2 = x_3 \}$ and also $\hat{\mu}_{\texttt{diff}_u(X,Y)}(E_2) = \hat{\mu}_{\texttt{diff}_u(X,Y)}(E_1)$.

**Theorem 1.** *Complete equi-propagation is uniformly stronger than unit propagation.*

*Proof.* Suppose formula $\varphi \models C$ where $C = (\ell_1 \vee \cdots \vee \ell_n)$ is a clause. Assume also that $E \in \mathcal{E}$ is such that $E \models \neg\ell_1, \ldots, E \models \neg\ell_{n-1}$. Unit propagation from $C \wedge E$ will infer $\ell_n$. Clearly $\varphi \wedge E \models (\ell_1 \vee \cdots \vee \ell_n) \wedge \neg\ell_1 \wedge \cdots \wedge \neg\ell_{n-1} \models \ell_n = 1$ and hence $\{\ell_n = 1\} \in \hat{\mu}_\varphi(E)$. Thus complete equi-propagation will infer everything inferred by unit propagation for *any* clausal representation of $\varphi$.

**Boolean Unifiers** It is convenient to view equi-formula in a generic "solved-form" as a substitution, $\theta_E$, which is a (most general) unifier for the equations in $E$. Boolean substitutions generalize assignments in that variables can be bound also to literals. A Boolean *substitution* is an idempotent mapping $\theta : \mathcal{B} \to \mathcal{L}_{0,1}$ such that $dom(\theta) = \{ b \in B \,|\, \theta(b) \neq b \}$ is finite and $\forall.b \in \mathcal{B}.\ \theta(b) \neq \neg b$. It is viewed as the set $\theta = \{ b \mapsto \theta(b) \,|\, b \in dom(\theta) \}$. We can apply $\theta$ to another substitution $\theta'$, to obtain substitution $(\theta \cdot \theta') = \{ b \mapsto \theta(\theta'(b)) \,|\, b \in dom(\theta) \cup dom(\theta') \}$. A *unifier* for equi-formula $E$ is a substitution $\theta$ such that $\models \theta(e)$, for each $e \in E$. A *most-general unifier* for $E$ is a substitution $\theta$ such that for any unifier $\theta'$ of $E$, there exists substitution $\gamma$ where $\theta' = \gamma \cdot \theta$.

*Example 2.* Consider the equi-formula $E \equiv \{b_1 = \neg b_2, \neg b_3 = \neg b_4, b_5 = b_6, b_6 = b_4, b_7 = 1, b_8 = \neg b_7\}$ then a unifier $\theta$ for $E$ is $\{b_2 \mapsto \neg b_1, b_4 \mapsto b_3, b_5 \mapsto b_3, b_6 \mapsto b_3, b_7 \mapsto 1, b_8 \mapsto 0\}$. Note that $\theta(E)$ is the trivially true equi-formula $\{b_1 = \neg\neg b_1, \neg b_3 = \neg b_3, b_3 = b_3, b_3 = b_3, 1 = 1, 0 = \neg 1\}$.

Let $\prec$ be a total (strict) order on $\mathcal{B}$, extended to an order on $\mathcal{L}_{0,1}$ such that $0 \prec 1$ and $\forall.b \in B$, $1 \prec b$ and $b \approx \neg b$. We define a canonical most-general unifier for any satisfiable equi-formula $E$: $\texttt{unify}_E = \lambda b.\min \{ \ell \in \mathcal{L}_{0,1} \,|\, E \models b = \ell \}$. We can compute $\texttt{unify}_E$ in almost linear (amortized) time using a variation of the union-find algorithm [20].

*Example 3.* For the equi-formula $E$ and substitution $\theta$ from Example 2 we have that $\texttt{unify}_E = \theta$ where the ordering is $0 \prec 1 \prec b_1 \prec b_2 \prec \cdots \prec b_8$.

The following allows us to replace formula $\varphi$ by $\texttt{unify}_E(\varphi)$, and provides an alternative, more efficient to implement, definition for complete equi-propagation.

**Proposition 1.**
$\varphi \wedge E \leftrightarrow \texttt{unify}_E(\varphi) \wedge E$

**Proposition 2.**
$\hat{\mu}_\varphi(E) \leftrightarrow E \wedge \{e \in \mathcal{E} \mid \texttt{unify}_E(\varphi) \models e\}$

Before presenting the proofs of Propositions 1 and 2 we present two additional propositions. First some notation: The projection of variable $b$ from $\varphi$ is the formula $(\exists b.\varphi)$ defined by $\varphi[b] \vee \varphi[\neg b]$. The projection $(\exists B.\varphi)$ for a set of variables $B \subseteq \mathcal{B}$, is defined in the natural way.

**Proposition 3.** *Let $\varphi$ be a propositional formula and $E$ an equi-formula. Then, $\exists dom(\texttt{unify}_E). (E \wedge \varphi) \leftrightarrow \texttt{unify}_E(\varphi)$.*

*Proof.* Let $A$ be a satisfying assignment for $\texttt{unify}_E(\varphi)$. Since $\texttt{unify}_E(\varphi)$ does not involve variables in $dom(\texttt{unify}_E)$ we can extend $A$ to

$$A' = A \cup \{ b \mapsto A(\texttt{unify}_E(b)) \,|\, b \in dom(\texttt{unify}_E) \}$$

By construction, $A'$ is a satisfying assignment of $E \wedge \varphi$ because $\forall \ell = \ell' \in E$, $A'(\ell) = A'(\ell')$, and $A'(\varphi) = A \cdot \texttt{unify}_E(\varphi)$ is a tautology, and hence $A' \models \varphi$. So, $A$ is a satisfying assignment for $E \wedge \varphi$. Let $A'$ be a satisfying assignment for $E \wedge \varphi$. Then $A'$ is a unifier of $E$. Hence, $\forall \ell = \ell' \in E$, $A'(\ell) = A'(\ell')$ which implies that $\forall b \mapsto \ell \in \texttt{unify}_E$, $A'(b) = A'(\ell)$. Let $A = \{ b \mapsto \ell \in A' \,|\, b \notin dom(\texttt{unify}_E) \}$. Now $A(\texttt{unify}_E(\varphi)) = A'(\varphi)$ by construction and $A'(\varphi)$ is a tautology. Hence $A$ is a satisfying assignment of $\texttt{unify}_E(\varphi)$. □

**Proposition 4.** *Let $\varphi$ be a formula, and $E$ a set of Boolean equalities, and $e$ a Boolean equality $\varphi \wedge E \models e$ iff $\texttt{unify}_E(\varphi) \models \texttt{unify}_E(e)$.*

*Proof.* ($\Leftarrow$) Suppose $\texttt{unify}_E(\varphi) \not\models \texttt{unify}_E(e)$ then there is a satisfying assignment $A$ of $\texttt{unify}_E(\varphi)$ where $A(\texttt{unify}_E(e))$ is *false*. By Proposition 3, $A$ is also a model of $\exists dom(\texttt{unify}_E).E \wedge \varphi$ and hence can be extended to a solution $A'$ of $E \wedge \varphi$. But $A'(e) \equiv A(\texttt{unify}_E(e))$ by definition, hence $E \wedge \varphi \not\models e$.
($\Rightarrow$) Suppose $\varphi \wedge E \not\models e$ then there is a satisfying assignment $A$ of $\varphi \wedge E$ where $A(e)$ is *false*. $A \models E \wedge \varphi$ then $A$ models $\texttt{unify}_E(\varphi)$. Similarly since $A$ is a model of $E$ $A(e) \equiv A(\texttt{unify}_E(e))$, hence $\texttt{unify}_E(\varphi) \not\models \texttt{unify}_E(e)$. □

*Proof (of Proposition 1).* Clearly $\varphi \wedge E \rightarrow \exists dom(\texttt{unify}_E).\varphi \wedge E \leftrightarrow \texttt{unify}_E(\varphi)$ by Proposition 3 and hence $\varphi \wedge E \rightarrow \texttt{unify}_E(\varphi) \wedge E$. For the reverse implication, let $A$ be a solution of $\texttt{unify}_E(\varphi) \wedge E$, then it is a unifier of $E$ and hence $A(b) = A(l)$ for each $b \mapsto l \in \texttt{unify}_E$ Then $A \models \varphi$ since $A \models \texttt{unify}_E(\varphi)$ and these two only differ by replacing $b$ with $l$ where $b \mapsto l \in \texttt{unify}_E$. Hence $\texttt{unify}_E(\varphi) \wedge E \rightarrow \varphi \wedge E$.
□

*Proof (of Proposition 2).* Let $E' = E \wedge \{e \in \mathcal{E} \mid \texttt{unify}_E(\varphi) \models e\}$. Clearly $\hat{\mu}_\varphi(E) \rightarrow E$ and $\hat{\mu}_\varphi(E) \rightarrow e$ where $\texttt{unify}_E(\varphi) \rightarrow e$ since $E \wedge \varphi \rightarrow \texttt{unify}_E(\varphi)$ by Proposition 3. Hence $\hat{\mu}_\varphi(E) \rightarrow E'$. Consider $e \in \hat{\mu}_\varphi(E)$. Then $\varphi \wedge E \rightarrow e$. Hence $\texttt{unify}_E(\varphi) \rightarrow \texttt{unify}_E(e)$ by Proposition 4, so $\texttt{unify}_E(e) \in E'$ Finally $E \wedge \texttt{unify}_E(e) \leftrightarrow E \wedge e$ using Proposition 1 and hence $E' \rightarrow e$. □

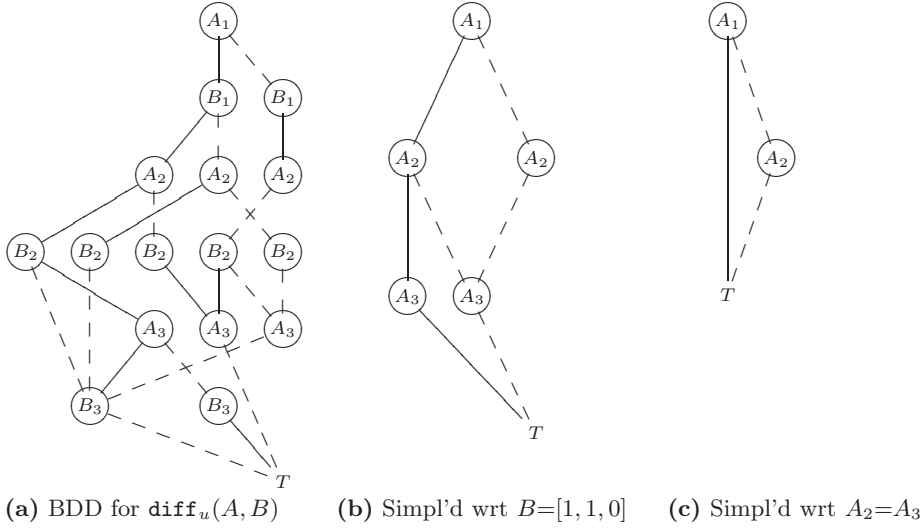

**(a)** BDD for $\text{diff}_u(A, B)$    **(b)** Simpl'd wrt $B=[1,1,0]$    **(c)** Simpl'd wrt $A_2=A_3$

**Fig. 1.** BDDs for the formula (a) $\varphi \equiv \text{unary}_3(A, [0,3]) \wedge \text{unary}_3(B, [0,3]) \wedge \text{diff}(A,B)$ (b) $\text{unify}_E(\varphi)$ where $E = \{B_1 = 1, B_2 = 1, B_3 = 0\}$ and (c) $\text{unify}_{E'}(\varphi)$ where $E' = E \cup \{A_2 = A_3\}$. The full (dashed) line corresponds to the true (false) child. Edges to the target "F" are omitted for brevity.

**Implementing complete equi-propagators** A complete equi-propagator is straightforward to implement using binary decision diagrams (BDDs). Consider Boolean formula $\varphi$ and equi-formula $E$. Then, for equation $(\ell_1 = \ell_2)$, based on Proposition 2, we can test the condition, $\text{unify}_E(\varphi) \models (\ell_1 \leftrightarrow \ell_2)$ using a standard BDD containment test e.g., "bddLeq" in [18]. This test can be performed for all relevant equations involving variables from $\text{unify}_E(\varphi)$ (and constants 0,1).

*Example 4.* Consider the BDD shown in Figure 3(a) which represents the formula: $\varphi \equiv \text{unary}_3(A, [0,3]) \wedge \text{unary}_3(B, [0,3]) \wedge \text{diff}(A,B)$. Suppose that $E$ is $\{B_1 = 1, B_2 = 1, B_3 = 0\}$. The BDD for $\text{unify}_E(\varphi)$ is shown in Figure 3(b). It is easy to see from the BDD that equi-propagation determines that $\text{unify}_E(\varphi) \models A_2 = A_3$. Indeed $\hat{\mu}_\varphi(E) = E' = E \cup \{A_2 = A_3\}$.

We apply complete equi-propagation in cases when BDDs are guaranteed to be polynomial in the size of the constraints they propagate for. The following result holds for an arbitrary constraint $\varphi$, so it also holds for $\text{unify}_E(\varphi)$.

**Proposition 5.** *Let $c(Xs)$ be an arbitrary constraint about integer variables $Xs = [X_1, \ldots, X_k]$ each represented with $n$ bits in the order encoding. Then, the number of nodes in the BDD representing $c(Xs)$ is bound by $O(n^k)$.*

*Proof.* (Sketch) There are only $n + 1$ legitimate states for each $n$ bit unary variable, and the BDD cannot have more nodes than possible states. □

**Implementing adhoc equi-propagators** Most simple constraints have a fixed small arity and hence complete equi-propagators using BDD are polynomially bounded. However, this is not the case for global constraints where the arity is not fixed. In this case we can define an adhoc, possibly incomplete, equi-propagator. We demonstrate this for the `allDiff`($[U_1, \ldots, U_m]$) constraint where each $U_i$ is represented in $n$ bits.

*Example 5.* Consider $Us = [U_1, \ldots, U_5]$ where the $U_i = \langle x_{i1}, \ldots, x_{i9} \rangle$ are integer variables in the range $[0, 9]$. Given $E$, we denote $\text{unify}_E(Us) = [U'_1, \ldots, U'_5]$ and illustrate equi-propagator $\mu_\varphi(E) = E \cup E'$ for $\varphi = \text{allDiff}_u(Us)$: **(1)** Consider $E = \{x_{12}=1, x_{13}=0\}$. Denoting $E_a = \{x_{1j} = j \leq 2 \,|\, 1 \leq j \leq 9\}$ (e.g. $U'_1 = 2$), and $E_b = \{x_{i2} = x_{i3} \,|\, 2 \leq i \leq 5\}$ (e.g. $U'_i \neq 2$ for $i > 1$), the propagator adds $E' = E_a \cup E_b$. **(2)** Consider $E = E_b \cup E_c$ where $E_c = \{x_{i5} = 0 \,|\, 1 \leq i \leq 5\}$ (e.g. $U_i \leq 4$) and $E_b$ is from the previous case. In this case, only $U_1$ can take the value 2. So a propagator adds equations imposing that $U'_1 = \langle 1, 1, 0, \ldots, 0 \rangle$. **(3)** Consider $E = E_c \cup E_d$ where $E_d = \bigcup \{x_{i1} = x_{i2}, x_{i3} = x_{i4} \,|\, 3 \leq i \leq 5\}$ (e.g. only $U_1$ and $U_2$ can take the values 1 and 3) and $E_c$ is from the previous case. A propagator adds $E' = \cup \{x_{i1} = 1, x_{i2} = x_{i3}, x_{i4} = 0 \,|\, i \in \{1, 2\}\}$.

The following is essentially the usual domain consistent propagator for the `allDiff` constraint [17] applied to the unary encoding.

**Definition 1 (ad-hoc equi-propagator for `allDiff`).** An equi-propagator for $\varphi = \text{allDiff}_u(Us)$ where $Us = [U_1, \ldots, U_m]$ is defined as $\mu_\varphi(E) = E \cup E'$ where $E' = \{U'_i(v) = U'_i(v+1) \,|\, i \in \{1, \ldots, m\} - H, v \in V\}$ if there exists a Hall set $H \subseteq \{1, \ldots, m\}$ where $V = \cup_{i \in H} dom(U'_i)$, $|V| = |H|$ and denoting $\text{unify}_E(Us) = [U'_1, \ldots, U'_m]$. Otherwise, $E' = \emptyset$.

After a Hall set $H$ is detected (and equi-propagation has triggered), we also apply an additional decomposition rule:

$$\text{allDiff}(Us) \mapsto \text{allDiff}([U_i \mid i \in H]) \wedge \text{allDiff}([U_i \mid i \in \{1, \ldots, m\} - H])$$

The benefit arises because the first `allDiff` constraint is guaranteed to represent a permutation which then benefits from $\text{permutation}_{\#}$.

For the three cases in Example 5 we have: **(1)** $H=\{1\}$ and $V=\{2\}$, **(2)** $H=\{2,3,4,5\}$, and $V=\{0,1,3,4\}$, and **(3)** $H=\{3,4,5\}$ and $V=\{0,2,4\}$. Indeed we can convert any finite domain propagator to an equi-propagator. The following holds simply because the unary encoding can represent arbitrary domains.

**Proposition 6.** *Let $E \in \mathcal{E}$ and $c(Xs)$ be a constraint over integer variables $Xs = [X_1, \ldots, X_m]$. Let $\text{unify}_E(Xs) = [X'_1, \ldots, X'_m]$. Suppose $D$ is the mapping from variables to sets of value $D(X_i) = dom(X'_i)$ and suppose propagator $f$ for $c(Xs)$ maps $D$ to $D'$. Then a correct equi-propagator for $c(Xs)$ discovers new equality literals $E' = \{X'_i(v) = X'_i(v+1) \mid i \in \{1, \ldots, m\}, v \in D(X_i) - D'(X_i)\}$.*

Note that complete equi-propagators can determine more information than finite domain propagation as illustrated by the example for $E_2$ in Example 1. To complete this section, consider the following example.

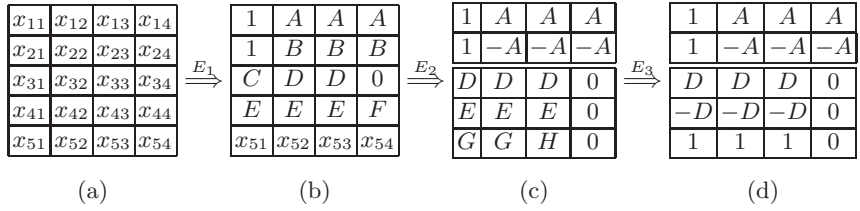

**Fig. 2.** An example of equi-propagation with $E_1$ to specify that $X_1 \in \{1,4\}$, $X_2 \in \{1,4\}$, $X_3 \in \{0,1,3\}$, $X_4 \in \{0,3,4\}$, and $X_5 \in \{0,1,2,3,4\}$.

*Example 6.* Consider a constraint $\texttt{allDiff}(Xs)$ where $Xs = [X_1, \ldots, X_5]$ with each $X_i$ in the interval $[0,4]$, depicted as Fig. 2(a). Consider also the equi-formula $E_1$ which specifies that $X_1, X_2 \in \{1,4\}$, $X_3 \in \{0,1,3\}$, $X_4 \in \{0,3,4\}$. Fig. 2(b) depicts $\texttt{unify}_{E_1}(Xs)$. Constraint simplification proceeds in two steps. First, equi-propagation adds equi-formula $E_2$, the affect of which is depicted as Fig. 2(c) where the constraint is also decomposed to constraints: The upper part, $\texttt{allDiff}(X_1, X_2)$, and the lower part $\texttt{allDiff}(X_3, X_4, X_5)$. In the second step equi-propagation adds equi-formula $E_3$, the impact of which is depicted as Fig. 2(d). The original constraint is now fully solved, $Xs$ is represented using only 2 propositional variables and the CNF encoding will contain no clauses.

## 4 Optimized SAT encodings using Equi-propagation

Boolean equi-propagation is at the foundation of our optimizing CNF compiler. The compiler repeatedly applies: equi-propagation, constraint decomposition, restriction and elimination, and finally outputs CNF encodings. We assume that each constraint comes with an associated equi-propagator.

Given a conjunction $\Phi$ of constraints, we first apply equi-propagators. Each such application effectively removes at least one bit from the Boolean representation of $\Phi$. During this process, when no further equi-propagators can be applied, we may apply a decomposition rule to a high-level constraint, introducing additional low-level constraints, but without introducing additional bits in the system. The actual implementation is of course less naive. It takes care to wakeup equi-propagators only when they may generate new information, and it makes use of the most efficient implementation of the equi-propagator possible, so avoiding BDD based propagators if we have an equivalent propagator implemented directly.

The complexity of the compiler is measured in the size of the constraint system $\Phi$ it is optimizing. Denote by $|\Phi|_c$, the total number of low-level constraints in $\Phi$ after decomposing all high-level constraints, and by $|\Phi|_b$ the total number of Boolean variables in the bit representation of $\Phi$. Assuming that equi-propagators are of polynomial cost, then so is the cost of running the compiler itself.

**Proposition 7.** *Let $\Phi$ be a conjunction of high- and low-level (finite domain) constraints. Then the number of equi-propagation steps performed when compiling $\Phi$ is bound by $O(|\Phi|_c \times |\Phi|_b)$.*

*Proof.* (sketch) Each pass of the algorithm covers at most $|\Phi|_c$ constraints and removes at least one of the $|\Phi|_b$ Boolean variables from $\Phi$. □

After equi-propagation and constraint decomposition triggers no more, we apply constraint restriction and elimination rules. We say that a constraint $\varphi$ is redundant with respect to an equi-formula $E$ if either (a) $\text{unify}_E(\varphi)$ is a tautology or (b) there exists another constraint $\varphi'$ in the constraint store such that $\text{unify}_E(\varphi') \models \text{unify}_E(\varphi)$. Our implementation is tuned to identify a collection of adhoc cases. However in general, where BDDs have been applied to implement complete propagators, such tests are particularly easy. Testing for (a) is trivial. Testing for (b) is also straightforward for BDDs e.g., using "`bddLeq`" in [18]. However we only apply this rule in a restricted form due to the quadratic time-complexity of examining all pairs of constraints. Namely, to determine cases of the form $true_u(X_1, \ldots, X_n) \models c(X_1, \ldots, X_n)$ where the constraint is redundant with respect to the unary encoding of its variables.

*Example 7.* Take $\varphi = \text{diff}_u(A, B)$ and $E'$ from Example 4. The BDD for $\varphi'' = \text{unify}_{E'}(\varphi)$ is shown in Figure 3(c). One can check that $\text{unary}(A, [0, 4]) \models \varphi''$ using "bddLeq" indicating that the original constraint $\text{diff}(A, B)$ is redundant.

In the final stage, when no further simplification applies, constraints are encoded to CNF formula. This can be performed either using their Boolean specification, or if BDD based propagators were applied, then we can read off the encoding from the BDD using standard techniques.

Redundant constraints (with subscript # in the name) that were introduced in the model only to improve equi-propagation need not be encoded to CNF clauses. However, when we expect such redundant clauses to facilitate unit propagation during SAT solving, then we do add them. For instance, we add clauses to encode redundant $\text{permutation}_\#$ constraints. Each such constraint $\varphi' = \text{permutation}_\#([U_1, \ldots, U_m])$ is affiliated with a corresponding $\text{allDiff}$ constraint. If $S = \cup_{i=1}^m dom(U_i)$, $|S| \neq m$ then the $\text{allDiff}$ constraint does not represent a permutation and nothing is added. Otherwise we create additional Boolean variables $b_{iv}$ to represent the expressions $U_i = v, v \in S$. Let the unary encoding of $U_i$ be $\langle u_1, \ldots, u_k \rangle$. We add clauses encoding $b_{iv} \leftrightarrow (u_v \wedge \neg u_{v+1})$ to connect these to the unary encoding, and the clauses $\vee_{i=1}^m b_{iv}, \forall v \in S$ to get better propagation from permutations.

## 5 Implementation, Experiments, and Extensions

All experiments were performed on an Intel Core 2 Duo E8400 3.00GHz CPU with 4GB memory under Linux (Ubuntu lucid, kernel 2.6.32-24-generic).[5] Our

---

[5] The benchmark instances and encodings can be viewed at
http://www.cs.bgu.ac.il/~amitmet/csp2011/.

prototype constraint compiler is written in Prolog and run using SWI Prolog
v5.10.2 64-bits. Complete propagators are implemented using the BDD package,
CUDD v2.4.2. Comparisons with Sugar (v1.14.7) are based on the use of identical
constraint models, apply the same SAT solver (CryptoMiniSat v2.5.1), and run
on the same machine. Comparisons with Minion (v0.10) are based on the use of
identical constraint models, and run on the same machine (with minor differences
due to syntax). For each of the example problems we extend (our description of)
the constraint modelling language as required for the benchmarks.

| instance | | compiler | | | Sugar | | 3D | | CSP'08 | OSC'09 | FS'09 |
|---|---|---|---|---|---|---|---|---|---|---|---|
| num | un/sat | compl (sec.) | cnf size (clauses) | SAT (sec.) | cnf size (clauses) | SAT (sec.) | cnf size (clauses) | SAT (sec.) | (sec.) | (sec.) | (sec.) |
| 1 | sat | 0.41 | 6509 | 2.46 | 140315 | 37.36 | 6507 | 0.09 | 31.55 | 34.81 | 6.44 |
| 2 | sat | 0.33 | 7475 | 0.02 | 140920 | 234.70 | 7438 | 0.74 | 137.60 | 99.84 | 44.80 |
| 3 | sat | 0.38 | 6531 | 0.02 | 140714 | 17.02 | 6512 | 0.02 | 10.57 | 12.25 | 2.53 |
| 4 | sat | 0.38 | 6818 | 0.61 | 141581 | 90.64 | 6811 | 0.08 | 47.24 | 273.36 | 157.58 |
| 5 | sat | 0.35 | 7082 | 0.32 | 140431 | 206.03 | 7099 | 0.14 | 27.33 | 24.87 | 22.30 |
| 6 | sat | 0.33 | 7055 | 0.45 | 140625 | 67.84 | 7044 | 1.11 | 35.78 | 108.60 | 12.58 |
| 7 | sat | 0.33 | 7711 | 2.36 | 142200 | 60.97 | 7684 | 0.08 | 57.23 | 67.32 | 341.62 |
| 8 | sat | 0.35 | 7426 | 0.05 | 140784 | 34.43 | 7367 | 0.04 | 43.88 | 1.52 | 6.08 |
| 9 | sat | 0.37 | 6602 | 0.28 | 137589 | 33.76 | 6609 | 0.41 | 25.15 | 9.52 | 3.01 |
| 10 | sat | 0.36 | 6784 | 0.17 | 142303 | 50.86 | 6799 | 0.06 | 26.16 | 27.80 | 12.66 |
| 11 | unsat | 0.45 | 6491 | 0.05 | 140603 | 39.02 | 6534 | 0.03 | 19.47 | 30.92 | 5.30 |
| 12 | unsat | 0.23 | 1 | 0.00 | 139037 | 0.58 | 7393 | 0.00 | 0.36 | 0.05 | 0.81 |
| 13 | unsat | 0.28 | 1 | 0.00 | 141295 | 0.90 | 6555 | 0.00 | 1.47 | 0.16 | 0.80 |
| 14 | unsat | 0.28 | 1 | 0.00 | 140706 | 2.25 | 7173 | 0.00 | 1.40 | 0.29 | 0.80 |
| 15 | unsat | 0.38 | 6063 | 0.05 | 140224 | 35.93 | 6104 | 0.06 | 32.39 | 58.41 | 4.77 |

**Table 1.** QCP results for $25 \times 25$ instances with 264 holes

**Quasigroup Completion Problems** (QCP) are given as an $n \times n$ board of
integer variables (in the range $[1, n]$) in which some are assigned integer values.
The task is to complete the board, assigning values to all variables, so that no
column or row contains the same value twice. A model for a QCP instance is a
conjunction of `allDiff` constraints corresponding to the variables (and values)
in its rows and columns.

Table 1 illustrates results for 15 (of the largest) instances from the 2008 CSP
competition[6] with data for our compiler (compilation time, number of clauses,
SAT solving time), Sugar (number of clauses, subsequent SAT solving time), the
so-called 3D SAT encoding of [12] (number of clauses after unit propagation, SAT
solving time), and from: CSP'08 (the winning result from the 2008 competition),
OSC'09 and FS'09 (results for lazy clause generation solvers reported in [15] and
[9]). It is, by now, accepted that the 3D encoding is strong for QCP problems, a
fact echoed by the results of Table 1. Observe that for 3 instances, unsatisfiable
is detected directly by the compiler (where the CNF contains 1 empty clause).

Table 2 illustrates results for larger ($40 \times 40$, satisfiable) instances[7] with 800-
1000 holes. We compare the order-encoding (compiled) and the 3D-encoding
(with unit propagation). The CNF sizes before compilation/unit propagation are circa 2.74 million clauses for the order-encoding and 3.74 for the 3D-encoding. The advantage of the 3D encoding is no longer clear.

---

[6] http://www.cril.univ-artois.fr/CPAI08/
[7] Generated using lsencode from http://www.cs.cornell.edu/gomes/SOFT.

| inst. | order enc. | | 3D enc. | | inst. | order enc. | | 3D enc. | |
|---|---|---|---|---|---|---|---|---|---|
| 800 | CNF | SAT | CNF | SAT | 1000 | CNF | SAT | CNF | SAT |
| holes | mCl | (sec) | mCl | (sec) | holes | mCl | (sec) | mCl | (sec) |
| 1 | 0.11 | 18.24 | 0.13 | 6.87 | 1 | 0.31 | 0.40 | 0.38 | 27.78 |
| 2 | 0.11 | 2.88 | 0.13 | 3.70 | 2 | 0.31 | 0.39 | 0.38 | 0.33 |
| 3 | 0.11 | 6.54 | 0.13 | 2.50 | 3 | 0.31 | 0.39 | 0.39 | 19.76 |
| 4 | 0.11 | 0.34 | 0.13 | 1.47 | 4 | 0.31 | 0.39 | 0.38 | 8.73 |
| 5 | 0.11 | 21.50 | 0.13 | 7.09 | 5 | 0.31 | 0.39 | 0.38 | 0.35 |
| 6 | 0.11 | 0.68 | 0.13 | 1.78 | 6 | 0.30 | 0.39 | 0.37 | 3.13 |
| 7 | 0.11 | 13.79 | 0.13 | 2.75 | 7 | 0.30 | 9.22 | 0.37 | 0.34 |
| 8 | 0.11 | 25.16 | 0.13 | 0.48 | 8 | 0.32 | 0.39 | 0.39 | 0.36 |
| 9 | 0.11 | 9.46 | 0.13 | 12.92 | 9 | 0.31 | 0.38 | 0.38 | 4.21 |
| 10 | 0.11 | 4.59 | 0.13 | 1.43 | 10 | 0.30 | 0.65 | 0.37 | 8.26 |

**Table 2.** QCP $40 \times 40$. CNF size in **million's** clauses

**Nonogram Problems** are expressed as a board of cells to color black or white, given clues per row and column of the board. A clue is a number sequence indicating blocks of cells to be colored black. For example, the clue $\langle 4, 8, 3\rangle$ on a row (or column) indicates that it should contain contiguous blocks of 4, 8 and 3 black cells (in that order) and separated by non-empty sequences of white cells.

A Nonogram puzzle is modeled as a Boolean matrix with constraints per row and column, each about a clue (sequence of numbers) $\langle b_1, \ldots, b_k \rangle$, and about a Boolean vector, $Vec$ (a row or column of the matrix). Each number $b_i$ is associated with an integer variable indicating the index in $Vec$ where block $b_i$ starts. For notation, if $U = \langle u_1, \ldots, u_n \rangle$ is an integer variable (order-encoding) then $U^{+c}$ is the vector with $c$ ones prefixing the bits of $U$ and represents the value $U + c$. Similarly, if $U$ is greater than $c$ then $U^{-c} = \langle u_{c+1}, \ldots, u_n \rangle$ represents the value $U - c$. We introduce two additional constraints

| instance | | compiler | | | | |
|---|---|---|---|---|---|---|
| id | size | comp | cnf | sat | BGU | Walt. |
| 9717 | (30x30) | 0.13 | 14496 | 124.43 | $\infty$ | $\infty$ |
| 10000 | (50x40) | 0.28 | 44336 | 40.66 | $\infty$ | $\infty$ |
| 9892 | (40x50) | 0.57 | 30980 | 0.44 | $\infty$ | $\infty$ |
| 2556 | (45x65) | 0.13 | 2870 | 0.00 | 15.85 | 0.4 |
| 10088 | (63x52) | 0.64 | 78482 | 1.26 | 0.27 | 0.08 |
| 2712 | (47x47) | 0.31 | 43350 | 0.92 | 5.98 | 4.95 |
| 8478 | (50x50) | 0.40 | 51027 | 0.95 | 0.89 | $\infty$ |
| 6727 | (80x80) | 1.11 | 156138 | 2.86 | 0.5 | 0.17 |
| 8098 | (19x19) | 0.02 | 3296 | 0.06 | 209.54 | 8.63 |
| 6574 | (25x25) | 0.10 | 7426 | 0.03 | 37.56 | 2.94 |

**Table 3.** Human Nonograms Results

$\boxed{4}$ $\texttt{block}(U_1, U_2, Vec)$ $\qquad\qquad$ $\boxed{5}$ $\texttt{leq}(U_1, U_2)$

The first specifies that for a bit vector $Vec$ the variables in the indices greater than value $U_1$ and less equal value $U_2$ (with $U_1 \leq U_2$) are true. The second specifies that for integer variables $U_1$ and $U_2$ in the order-encoding, $U_1 \leq U_2$. The Boolean functions corresponding to constraints of these forms are as follows:

$$\begin{array}{l} \texttt{block}(U_1, U_2, \langle x_1, \ldots, x_n \rangle) = \bigwedge_{i=1}^{n} (\neg U_1(i) \wedge U_2(i) \rightarrow x_i \\ \texttt{leq}(\langle x_1, \ldots, x_n \rangle, \langle y_1, \ldots, y_n \rangle) = \bigwedge_{i=1}^{n} x_i \rightarrow y_i \end{array} \qquad (3)$$

*Example 8.* The constraints below model the position of block sequence $s = \langle 3, 1, 2 \rangle$ in $X = \langle x_1, \ldots, x_9 \rangle$. In the first column, integer variables, $U_1, U_2, U_3$ model the start positions of the three blocks. In the second column, the start position of a block is required to be at least one after the end position of its prede-cessor. In the third column, `block` constraints specify the black cells in the vector $X$, and in the fourth column the white cells in the block $\bar{X} = \langle\neg x_1, \ldots, \neg x_9\rangle$.

$\texttt{unary}(U_1,[1,9])$ $\phantom{\texttt{leq}(U_1^{+4},U_2)}$ $\texttt{block}(U_1^{-1}, U_1^{+3}, X)$ $\texttt{block}(0, U_1, \bar{X})$
$\texttt{unary}(U_2,[1,9])$ $\texttt{leq}(U_1^{+4},U_2)$ $\texttt{block}(U_2^{-1}, U_2^{+1}, X)$ $\texttt{block}(U_1^{+4}, U_2, \bar{X})$
$\texttt{unary}(U_3,[1,9])$ $\texttt{leq}(U_2^{+2},U_3)$ $\texttt{block}(U_3^{-1}, U_3^{+2}, X)$ $\texttt{block}(U_2^{+2}, U_3, \bar{X})$

Tables 3 & 4 compare ours to the two fastest documented Nonogram solvers: BGU (v1.0.2) [16] and Wolter (v1.09) [23]. Table 3 is about "human-designed" instances from [21]. These are the 10 hardest problems for the BGU solver. The first 8 puzzles have at least 2 solutions. The last 2 have a single solution.

| time (sec) | 0.20 | 0.50 | 1.00 | 10.00 | 30.00 | 60.00 |
|---|---|---|---|---|---|---|
| BGU | 279 | 3161 | 4871 | 4978 | 4989 | 4995 |
| Wolter | 4635 | 4782 | 4840 | 4952 | 4974 | 4976 |
| Compiler | 13 | 4878 | 4994 | 5000 | 5000 | 5000 |

Table 4. 5,000 Random Nonograms Results

Solving time is for determining the number of solutions (0, 1, or more). For our compiler, the columns indicate: compilation time, cnf size (number of clauses) and sat solving time. The final two columns are about the solution times for the BGU and Wolter solvers (running on the same machine). The timeout for these solvers (indicated by $\infty$) is 300 sec. Table 4 reports on a collection of 5,000 random puzzles from [22]. For each of the three solvers we indicate how many puzzles it solves within the given allocated time.

**BIBD Problems** (CSPlib problem 28) are defined by a 5-tuple of positive integers $\langle v, b, r, k, \lambda\rangle$ and require to partition $v$ distinct objects into $b$ blocks such that each block contains $k$ different objects, exactly $r$ objects occur in each block, and every two distinct objects occur in exactly $\lambda$ blocks. To model BIBD problems we introduce three additional constraints

$\boxed{6}$ $\texttt{sumBits}([B_1,\ldots,B_n],U)$ $\phantom{xxxx}$ $\boxed{7}$ $\texttt{uadder}(U_1,U_2,U_3)$

$\boxed{8}$ $\texttt{pairwise\_and}([A_1,\ldots,A_n],[B_1,\ldots,B_n],[C_1,\ldots,C_n])$

The first (high-level) constraint states that the sum of bits, $[B_1,\ldots,B_n]$ is the unary value $U$. It is defined by decomposition: split the bits into two parts, sum the parts, then add the resulting (unary) numbers. The sum of two unary numbers, $U_1 + U_2 = U_3$, is specified by the (low-level) constraint $\texttt{uadder}(U_1,U_2,U_3)$. To compute the scalar product of vectors $[A_1,\ldots,A_n]$ and $[B_1,\ldots,B_n]$ we use the $\texttt{pairwise\_and}$ constraint in combination with $\texttt{sumBits}$.

The model for BIBD instance $\langle v,b,r,k,\lambda\rangle$ is a Boolean incidence matrix with constraints: $\texttt{sumBits}(C,k)$ for each column $C$; $\texttt{sumBits}(R,r)$ for each row $R$; and for each pair of rows $R_i$, $R_j$ ($i < j$), $\texttt{pairwise\_and}(R_i,R_j,Vs)$ and $\texttt{sumBits}(Vs,\lambda)$. To break symmetry, we can reorder rows and columns of the matrix to assign fixed values in the first two rows and leftmost column: the first row contains $r$ ones, followed by zeros. The second row contains $\lambda$ ones, $r - \lambda$ zeros, $r - \lambda$ ones, and then zeros. The left column contains $k$ ones followed by zeros. This is the information that enables the compiler to simplify constraints.

Table 5 shows results comparing our compiler using the model we call SymB for symmetry breaking (compilation time, cnf size, and sat solving time) with the Minion constraint solver [11]. Ignore for now the last 3 columns about SatELite.

| instance | compiler (SymB) | | | Minion | | | SATELITE (SymB) | | |
|---|---|---|---|---|---|---|---|---|---|
| $\langle v, b, r, k, \lambda \rangle$ | comp | cnf size | SAT | [M'06] | SymB | SymB$^+$ | prepro | cnf size | SAT |
| | (sec.) | (clauses) | (sec.) | (sec.) | (sec.) | (sec.) | (sec.) | (clauses) | (sec.) |
| $\langle 7, 350, 150, 3, 50 \rangle$ | 1.34 | 494131 | 1.23 | 0.47 | 1.12 | 0.38 | 1.27 | 566191 | 1.65 |
| $\langle 7, 420, 180, 3, 60 \rangle$ | 1.65 | 698579 | 1.73 | 0.54 | 1.36 | 0.42 | 1.67 | 802576 | 2.18 |
| $\langle 7, 560, 240, 3, 80 \rangle$ | 3.73 | 1211941 | 13.60 | 0.66 | 1.77 | 0.52 | 2.73 | 1397188 | 5.18 |
| $\langle 8, 84, 42, 4, 18 \rangle$ | 0.25 | 64432 | 0.17 | 2.41 | $\infty$ | 0.70 | 0.31 | 73780 | 0.15 |
| $\langle 8, 98, 49, 4, 21 \rangle$ | 0.33 | 84993 | 0.23 | 5.63 | $\infty$ | 1.54 | 0.34 | 97588 | 0.33 |
| $\langle 12, 132, 33, 3, 6 \rangle$ | 0.95 | 180238 | 0.73 | 5.51 | $\infty$ | 1.76 | 1.18 | 184764 | 0.57 |
| $\langle 13, 26, 8, 4, 2 \rangle$ | 0.12 | 17570 | 0.05 | 5.46 | 0.47 | 0.16 | 0.22 | 17391 | 0.10 |
| $\langle 15, 45, 24, 8, 12 \rangle$ | 0.51 | 116016 | 8.46 | $\infty$ | $\infty$ | 75.87 | 0.64 | 134146 | $\infty$ |
| $\langle 15, 70, 14, 3, 2 \rangle$ | 0.56 | 81563 | 0.39 | 12.22 | 1.42 | 0.31 | 1.02 | 79542 | 0.20 |
| $\langle 16, 80, 15, 3, 2 \rangle$ | 0.81 | 109442 | 0.56 | 107.43 | 13.40 | 0.35 | 1.14 | 105242 | 0.35 |
| $\langle 19, 19, 9, 9, 4 \rangle$ | 0.23 | 39931 | 0.09 | 53.23 | 38.30 | 0.31 | 0.4 | 44714 | 0.09 |
| $\langle 19, 57, 9, 3, 1 \rangle$ | 0.34 | 113053 | 0.17 | $\infty$ | 1.71 | 0.35 | 10.45 | 111869 | 0.14 |
| $\langle 21, 21, 5, 5, 1 \rangle$ | 0.02 | 0 | 0.00 | 1.26 | 0.67 | 0.15 | 0.01 | 0 | 0.00 |
| $\langle 25, 25, 9, 9, 3 \rangle$ | 0.64 | 92059 | 1.33 | $\infty$ | $\infty$ | 0.92 | 1.01 | 97623 | 8.93 |
| $\langle 25, 30, 6, 5, 1 \rangle$ | 0.10 | 24594 | 0.06 | $\infty$ | 1.37 | 0.31 | 1.2 | 23828 | 0.05 |
| $\langle 31, 31, 6, 6, 1 \rangle$ | 0.08 | 8571 | 0.03 | $\infty$ | 2.10 | 0.36 | 0.28 | 8001 | 0.03 |
| Total | | | 40.53 | | | 84.40 | | | > 223.82 |

**Table 5.** BIBD results (180 sec. timeout)

We will come back to explain these in Section 6. All experiments were run on the same computer. We consider three different models for Minion: `[M'06]` indicates results using the BIBD model described in [11], `SymB` uses the same model we use for the SAT approach, `SymB`$^+$, is an enhanced symmetry breaking model with all of the tricks applied also in the `[M'06]` model. For the columns with no timeouts we show total times (for the compiler this includes compile time and sat solving). Note that by using a clever modeling of the problem we have improved also the previous runtimes for Minion.

**Word Design for DNA** (Problem 033 of CSPLib) seeks the largest parameter $n$, s.t. there exist a set $S$ of $n$ eight-letter words over the alphabet $\Sigma = \{A, C, G, T\}$ with the following properties: (1) Each word in $S$ has 4 symbols from $\{C, G\}$; (2) Each pair of distinct words in $S$ differ in at least 4 positions; and (3) For every $x, y \in S$: $x^R$ (the reverse of $x$) and $y^C$ (the word obtained by replacing each $A$ by $T$, each $C$ by $G$, and vice versa) differ in at least 4 positions.

In [10], the authors present the "*template-map*" strategy for this problem. Letters are modelled by pairs $\langle t_i, m_i \rangle$ of bits. For each eight-letter word, $\langle t_1, \ldots, t_8 \rangle$ is the *template* and $\langle m_1, \ldots, m_8 \rangle$ is the *map*. The authors pose conditions on a set of templates $T$ and a set of maps $M$ so that the cartesion product $S = T \times M$ will satisfy the requirements of the original problem. It is this template-map strategy that we model in our encoding. The authors report a solution composed from two template-maps $\langle T_1, M_1 \rangle$ and $\langle T_2, M_2 \rangle$ where $|T_1| = 6$, $|M_1| = 16$, $|T_2| = 2$, $|M_2| = 6$. This forms a set $S$ with $(6 \times 16) + (2 \times 6) = 108$ DNA words. Marc van Dongen reports a larger solution with 112 words. [8] To model this problem we introduce the two constraints (where $V_i$ are vectors of bits).

     9  `lexleq([V_1, ..., V_n])`        10  `lexleq(V_1, V_2)`

---

[8] See `http://www.cs.st-andrews.ac.uk/~ianm/CSPLib/`.

The first specifies that a list of vectors is ordered in the lexicographic order. It decomposes to the low-level constraint (the second) that specifies that a pair of vectors is ordered in the lexicographic order.

Using our compiler, we find a template and a map of sizes 14 and 8, the cartesian product of which gives a solution of size $14 \times 8 = 112$ words. The SAT solving time is less than 0.2 seconds. To show that there is no templete of size 15 and no map of size 9 takes 0.14 and 3.32 seconds respectively. This is a new result not obtainable using previous solving techniques. We obtain this result when symmetries are broken by ordering the vectors in $T$ and in $M$ lexicographically. Proving that there is no solution to the original DNA word problem with more than 112 words (not via the template-map strategy) is still an open problem.

## 6 Related Work and Conclusion

There is a considerable body of work on CNF simplification techniques with a clear trade-off between amount of reduction achieved and invested time. Most of these approaches determine binary clauses implied by the CNF, which is certainly enough to determine Boolean equalities. The problem is that determining all binary clauses implied by the CNF is prohibitive when the SAT model may involve many thousands of variables. Typically only some of the implied binary clauses are determined, such as those visible by unit propagation. The trade-off is regulated by the choice of the techniques applied to infer binary clauses, considering the power and cost. See for example [7] and the references therein.

In our approach, the beast is tamed by introducing a notion of locality. We do not consider the full CNF. Instead, by maintaining the original representation, a conjunction of constraints, each viewed as a Boolean formula, we can apply powerful reasoning techniques to separate parts of the model and maintain efficient preprocessing. Our specific choice, using BDD's for bounded sized formula, guarantees that reasoning is always polynomial in cost.

To illustrate one difference consider again Example 6 where equi-propagation simplifies the constraint so that it is expressed in 2 propositional variables and requires 0 clauses. In contrast, the CNF representing the `allDiff` constraint with the initial equations $E_1$ consists of 76 clauses with 23 variables and after applying SatELite [7] this is reduced to 57 clauses with 16 variables. Examining this reduced CNF reveals that it contains binary clauses corresponding to the equations in $E_2$ but not those from $E_3$.

Finally, we come back to (the last 3 columns in) Table 5 where a comparison with SatELite is presented. It is interesting to note that in some cases preprocessing results in smaller CNF and faster SAT solving, however in total (even if not counting the timeout for BIBD instance $\langle 15, 45, 24, 8, 12 \rangle$) equi-propagation is stronger.

Using equi-propagation on a high level view of the problem allows us to simplify the problem more aggressively than is possible with a CNF representation. The resulting CNF models can be significantly smaller than those result-ing from straight translation, and significantly faster to solve. Hence we believe that Boolean equi-propagation, combined with CNF simplification tools (such as SatELite), makes an important contribution to the encoding of CSPs to SAT.